\documentclass[11pt]{article}
\usepackage{amsmath}
\usepackage{amssymb}
\usepackage{amsthm}
\newcount\hour  \newcount\minutes  \hour=\time  \divide\hour by 60
\minutes=\hour  \multiply\minutes by -60  \advance\minutes by \time
\def\mmmddyyyy{\ifcase\month\or Jan\or Feb\or Mar\or Apr\or May\or Jun\or Jul\or
  Aug\or Sep\or Oct\or Nov\or Dec\fi \space\number\day, \number\year}
\def\hhmm{\ifnum\hour<10 0\fi\number\hour :%
  \ifnum\minutes<10 0\fi\number\minutes}

\newtheorem{theorem}{Theorem}[section]
  \newtheorem{definition}[theorem]{Definition}

\newcommand{\bigo}{{\protect\cal O}}
\newcommand{\calz}{{\protect\cal Z}}
\newcommand{\condition}{\,\mid \:}

\newcommand{\p}{\ensuremath{\mathrm{P}}}
\newcommand{\np}{\ensuremath{\mathrm{NP}}}
\newcommand{\numvars}{\ensuremath{\mathit{numvars}}}
\newcommand{\sat}{\ensuremath{\mathrm{SAT}}}
\newcommand{\conp}{\ensuremath{\mathrm{coNP}}}

\newcommand{\pairs}[1]{\mathopen\langle{#1}\mathclose\rangle}
\newcommand{\pair}[1]{{\mathopen\langle{#1}\mathclose\rangle}}

\newcommand{\npintconp}{\ensuremath{\np \cap \conp}}
\newcommand{\npinterconp}{\ensuremath{\npintconp}}
\newcommand{\npcapconp}{\ensuremath{\npintconp}}
\newcommand{\pisnotinter}{\ensuremath{\p \neq \npinterconp}}
\mathcode`\0="0030      %
\mathcode`\1="0031
\mathcode`\2="0032
\mathcode`\3="0033
\mathcode`\4="0034
\mathcode`\5="0035
\mathcode`\6="0036
\mathcode`\7="0037
\mathcode`\8="0038
\mathcode`\9="0039

\makeatletter
\def\@citex[#1]#2{\if@filesw\immediate\write\@auxout{\string\citation{#2}}\fi
  \def\@citea{}\@cite{\@for\@citeb:=#2\do
    {\@citea\def\@citea{,\linebreak[0]}\@ifundefined
       {b@\@citeb}{{\bf ?}\@warning
       {Citation `\@citeb' on page \thepage \space undefined}}%
\hbox{\csname b@\@citeb\endcsname}}}{#1}}
\newcommand{\singlespacing}{\let\CS=
\@currsize\renewcommand{\baselinestretch}{1}\tiny\CS}
\newcommand{\singlespacingplus}{\let\CS=
\@currsize\renewcommand{\baselinestretch}{1.25}\tiny\CS}
\newcommand{\doublespacing}{\let\CS=
\@currsize\renewcommand{\baselinestretch}{1.75}\tiny\CS}
\newcommand{\extradoublespacing}{\let\CS=
\@currsize\renewcommand{\baselinestretch}{1.9}\tiny\CS}
\newcommand{\nicenicespacing}{\let\CS=
\@currsize\renewcommand{\baselinestretch}{1.9}\tiny\CS}
\newcommand{\draftspacing}{\let\CS=
\@currsize\renewcommand{\baselinestretch}{2.0}\tiny\CS}
\newcommand{\hugedraftspacing}{\let\CS=
\@currsize\renewcommand{\baselinestretch}{2.4}\tiny\CS}
\newcommand{\niceohfivespacing}{\let\CS=\@currsize\renewcommand{\baselinestretch}{1.05}\tiny\CS}
\newcommand{\niceonespacing}{\let\CS=\@currsize\renewcommand{\baselinestretch}{1.1}\tiny\CS}
\newcommand{\nicetwospacing}{\let\CS=\@currsize\renewcommand{\baselinestretch}{1.2}\tiny\CS}
\newcommand{\nicethreespacing}{\let\CS=\@currsize\renewcommand{\baselinestretch}{1.3}\tiny\CS}
\newcommand{\singlespacingplusplus}{\let\CS=\@currsize\renewcommand{\baselinestretch}{1.35}\tiny\CS}
\newcommand{\nicefourspacing}{\let\CS=\@currsize\renewcommand{\baselinestretch}{1.4}\tiny\CS}
\newcommand{\nicefivespacing}{\let\CS=\@currsize\renewcommand{\baselinestretch}{1.5}\tiny\CS}
\newcommand{\nicesixspacing}{\let\CS=\@currsize\renewcommand{\baselinestretch}{1.6}\tiny\CS}
\newcommand{\nicesevenspacing}{\let\CS=\@currsize\renewcommand{\baselinestretch}{1.7}\tiny\CS}
\newcommand{\niceeightspacing}{\let\CS=\@currsize\renewcommand{\baselinestretch}{1.8}\tiny\CS}
\newcommand{\niceninespacing}{\let\CS=\@currsize\renewcommand{\baselinestretch}{1.9}\tiny\CS}
\makeatother%
\newcommand{\normalspacing}{\singlespacing}
\let\shortcite\cite
\let\citeyear\cite
\normalspacing

\begin{document}
\sloppy
\title{The Opacity of Backbones%
\thanks{%
A
preliminary
version of this 
paper appeared in
AAAI~2017~\cite{hem-nar:c:backbones-opacity}.}}
\author{
        Lane A. Hemaspaandra\\
        Department of Computer Science \\
        University of Rochester \\
        Rochester, NY 14627, USA
\and
David E. Narv\'{a}ez\\
College of Computing and Information Sciences\\
        Rochester Institute of Technology \\
        Rochester, NY 14623, USA}
      \date{January 14, 2019}
\maketitle

\begin{abstract}
  This paper approaches, using structural complexity theory, the question
  of whether there is a chasm between knowing an object exists and
  getting one's hands on the object or its properties.  In particular,
  we study the nontransparency of so-called backbones.
  A backbone of a boolean formula $F$ is a collection $S$ of its variables 
  for which there is a unique partial assignment $a_S$ such that
  $F[a_S]$ is
  satisfiable~\cite{kir-mon-sel-tro-zec:j:phase-transitions,gom-sel-wil:c:backdoors}.
  We show that, under the widely
  believed assumption that integer factoring is hard, there exist sets
  of boolean formulas that have obvious, nontrivial backbones yet
  finding the values, $a_S$, of those backbones is intractable.  We also
  show that, under the same assumption, there exist sets of boolean
  formulas that obviously have large backbones yet producing such a
  backbone $S$ is intractable.  
  Furthermore, we 
  show that if integer factoring is not merely worst-case hard but is 
  frequently hard, as is widely believed, 
  then the frequency of hardness in our two results
  is not too much less than that frequency.  These results hold
  more generally, namely, in the settings where, respectively,
  one's assumption is
  that $\p \neq \np \cap \conp$ or that some problem in $\np \cap \conp$
  is frequently hard.\\[5pt]Key Words: structural complexity theory, $\np \cap \conp$, backbones.
\end{abstract}

\section{Introduction}
An important concept in the study of the SAT problem is the notion of
backbones.  The term was first used 
by Monasson et
al.~\shortcite{kir-mon-sel-tro-zec:j:phase-transitions}, 
and the 
following formal definition was provided by 
Williams, Gomes, and
Selman~\shortcite{gom-sel-wil:c:backdoors}.
\begin{definition}\label{d:backbone} Let $F$ be a boolean formula.  
A collection 
$S$ of the variables of $F$ is said to be a
  \emph{backbone} if there is a unique partial assignment $a_S$
such that $F[a_S]$
  is satisfiable.
\end{definition}
In that definition, $a_S$ assigns a value (true or false) to each
variable in $S$, and $F[a_S]$ is a shorthand meaning $F$ except with
each variable in $S$ assigned the value specified for it in $a_S$.  
A
backbone $S$ is \emph{nontrivial} if $S \neq \emptyset$. 
The \emph{size} of a backbone $S$ 
is 
the number of variables 
in $S$.
For a
backbone $S$ (for formula $F$), we say that $a_S$ is the \emph{value}
of the backbone $S$.

For example, every satisfiable formula has the trivial backbone
$S = \emptyset$.  The formula $x_1 \land \overline{x_2}$ has four
backbones, $\emptyset$, $\{x_1\}$, $\{x_2\}$, and $\{x_1,x_2\}$, with
respectively the values (listing values as bit-vectors giving the
assignments in the lexicographical order of the names of the variables
in $S$) $\epsilon$, $1$, $0$, and $10$.  The formula
$x_1 \lor \overline{x_2}$ has no nontrivial backbones.  (Every 
formula that has a backbone will have a maximum backbone---a
backbone that every other backbone is a subset of.  Backbone variables
have been called ``frozen variables,'' because each of them is the
same over all satisfying assignments.)

As Williams, Gomes, and Selman~\shortcite{gom-sel-wil:c:backdoors}
note, ``backbone variables are useful in studying the properties of
the solution space of a...~problem.''\footnote{We mention in
passing 
that backbones also are very interesting for problems
even harder than SAT, such as 
model counting (see Gomes, Sabharwal, and Selman,~\citeyear{gom-sab-sel:b:model-counting-handbook-of-satisfiablity})---i.e., \#SAT~\cite{val:j:permanent}: counting
the number of satisfying assignments of a propositional boolean
formula---especially given the currently 
huge runtime differences 
between SAT-solvers and model counters.  After all, a backbone 
for a formula 
helpfully pinpoints into a particular 
subspace all the solutions that one is seeking to count.}

And that surely is so.  But it
is natural to hope to go beyond that and suspect that if formulas have
backbones, we can use those to help SAT solvers.  After all, if one is
seeking to get one's hands on a satisfying assignment of an $F$ that
has a backbone, one need but substitute in the value of the backbone
to have put all its variables to bed as to one's search, and thus to
``only'' have all the other variables to worry about.

The goal of the present paper is to understand, at least in a
theoretical sense, the difficulty of---the potential obstacles to
doing---what we just suggested.  We will argue that even for cases
when one can quickly (i.e., in polynomial time) recognize that a
formula has at least one nontrivial backbone, it can be intractable to
find one such backbone.  And we will argue that even for cases when
one can quickly (i.e., in polynomial time) find a large, nontrivial
backbone, it can be intractable to find the value of that backbone.
In particular, we will show that if integer factoring is hard, then
both the just-made claims hold.  Integer factoring is widely believed
to be hard; indeed, if it were in polynomial time, RSA 
(the Rivest-Shamir-Adleman cryptosystem) itself would
fall.  

In fact, integer factoring is even believed to be hard on average.
And we will be inspired by that to go beyond the strength of the
results mentioned above.  Regarding our results mentioned above, one
might worry that the ``intractability'' might be very infrequent,
i.e., merely a rare, worst-case behavior.  But we will argue that if
integer factoring---or indeed any problem in $\npintconp$---is
frequently hard, then the bad behavior types we mention above happen
``almost'' as often: If the frequency of hardness of integer factoring
is $d(n)$ for strings up to length $n$, then for some $\epsilon > 0$
the frequency of hardness of our problems is $d(n^\epsilon)$.

None of this means that backbones are not an excellent, important
concept.  Rather, this is saying---proving, in fact, assuming that 
integer factoring is as hard as is generally believed---that 
although the definition of
backbone is merely about a backbone existing, one needs to be
aware that going from a backbone existing to finding a backbone, and
going from having a backbone to knowing its value, can be 
computationally expensive challenges.

The following section presents our results,
and then the section after that, which we have placed after
the results section so that the reader is familiar with the results
and proofs before the related-work discussion, covers the related work.
After that is the conclusion and a technical appendix.

\section{Results}
Section~\ref{ss:basic} will formulate our results
without focusing on density.  Then in 
Section~\ref{ss:frequency}
we will discuss how the frequency of hardness of sets of the
type we have discussed is related to that of the sets in $\npinterconp$
having the 
highest frequencies of hardness.

\subsection{Core Results}\label{ss:basic}
We first look at whether there can be simple sets of formulas for
which one can easily compute/obtain a nontrivial backbone, yet one
cannot easily find the value of that backbone.  

Our basic result on this is stated below
as Theorem~\ref{t:hard-value}.
In this and most of our
results, we state as our hypothesis not that ``integer factoring cannot be 
done in polynomial time,'' but rather that ``$\p \neq \npinterconp$.''
This in fact makes our claims stronger ones than if they 
had as their hypotheses 
``integer factoring cannot be done in polynomial time,'' 
since it is well-known
(because the decision version of integer factorization is itself in
$\npinterconp$) that ``integer factoring cannot be done in polynomial time''
implies ``$\p \neq \npinterconp$.'' $\sat$ will, as usual, denote the 
set of satisfiable (propositional) boolean formulas.  (We do not 
assume that $\sat$ by definition is restricted to CNF formulas.)

\begin{theorem}\label{t:hard-value}
If $\p\neq \npintconp$, then there exists a set $A \in \p$, 
$A \subseteq \sat$, of 
boolean formulas such that:
\begin{enumerate}
\item There is a polynomial-time computable function $f$ such that 
$(\forall F\in A)[f(F)$ outputs a nontrivial backbone of $F]$.
\item There does not exist any polynomial-time computable function $g$ such
that $g(F)$ computes the value of backbone $f(F)$.
\end{enumerate}
\end{theorem}

Theorem~\ref{t:hard-value} remains true even if one restricts the 
backbones found by $f$ to be of size~$1$.  We state that, in a slightly 
more general form, as follows.  

\begin{theorem}\label{t:hard-value-size-k}
Let $k \in\{1,2,3,\ldots\}$. 
If $\p\neq \npintconp$, then there exists a set $A \in \p$, 
$A \subseteq \sat$, of 
boolean formulas such that:
\begin{enumerate}
\item There is a polynomial-time computable function $f$ such that 
$(\forall F\in A)[f(F)$ outputs a size-$k$ backbone of $F]$.
\item There does not exist any polynomial-time computable function $g$ such
that $g(F)$ computes the value of backbone $f(F)$.
\end{enumerate}
\end{theorem}

We defer the proof of Theorem~\ref{t:hard-value-size-k}
(which itself implies Theorem~\ref{t:hard-value})
until
after the statement of Theorem~\ref{t:hard-backbone}.

Now let us turn to the question of whether, when it is obvious that
there is at least one nontrivial backbone, it can be hard to
efficiently produce a nontrivial backbone.  The following 
theorem shows that, if integer factoring is hard, the answer is yes.

\begin{theorem}\label{t:hard-backbone}
If $\p\neq \npintconp$, then there exists a set $A \in \p$, 
$A \subseteq \sat$, of 
boolean formulas (each having at least one variable) such that:
\begin{enumerate}
\item Each formula $F\in A$ has a backbone whose size is at 
least 49\% of $F$'s total number of variables.
\item 
There does not exist any polynomial-time computable function $g$ such
that, on each $F \in A$, $g(F)$ outputs a backbone whose size is 
at least 49\%---or even at least 2\%---of $F$'s variables.
\end{enumerate}
\end{theorem}

\begin{proof}[Proof of
  Theorems~\ref{t:hard-value-size-k} and~\ref{t:hard-backbone}]
Since they share a proof structure,
we will prove Theorems~\ref{t:hard-value-size-k} and~\ref{t:hard-backbone}
hand-in-hand, and in a rather narrative fashion.
Each of those
theorems starts with the assumption that
$\p \neq \npcapconp$.  So let $B$ be some set
instantiating that, i.e., $B \in (\npcapconp) - \p$.  As all students
learn when learning that $\sat$ is NP-complete, we can efficiently
transform the question of whether a machine accepts a particular
string into a question about whether a certain boolean formula is
satisfiable~\cite{coo:c:theorem-proving,kar:b:reducibilities,lev:j:universal}.
The original work that did that did not 
require (and did not \emph{need} to require)
that the thus-created boolean
formula transparently revealed what machine and input had been the
input to the transformation.
But it was soon noted that one can ensure that the formula
mapped to transparently reveals the machine and input that were the
input to the transformation; see 
Galil~\shortcite{gal:j:encodings}
or our appendix.

Galil's insight can be summarized in the following strengthened
version of the standard claim regarding the so-called Cook-Karp-Levin 
Reduction.
Let $N_1$, $N_2$, $\ldots$ be a fixed, standard enumeration of clocked,
polynomial-time Turing machines,
and w.l.o.g.\ assume that $N_i$ runs
within time $n^i+i$ on inputs of length $n$, and that 
$N_i$ and $i$ are 
polynomially related in size and easily obtained from each other.
There is a function 
$r_{\textit{Galil-Cook}}$ (for conciseness, we are writing Galil-Cook 
rather than
Galil-Cook/Karp/Levin, although this version is closer to 
the setting 
of 
Karp and Levin than to that of Cook, 
since Cook used Turing reductions rather 
than many-one reductions) such that
\begin{enumerate}
\item For each $N_i$ and $x$: 
$x \in L(N_i)$ if and only if $r_{\textit{Galil-Cook}}(N_i,x) \in \sat$.
\item There is a polynomial $p$ such that 
$r_{\textit{Galil-Cook}}(N_i,x)$ runs within time 
polynomial (in particular, with $p$ being the polynomial) 
in
$|N_i|$ and $|x|^i+i$.
\item There is a polynomial-time function $s$ such that for each $N_i$ and $x$,
$s(r_{\textit{Galil-Cook}}(N_i,x))$ outputs the pair $(N_i,x)$.
\end{enumerate}

We will be using two separate applications of the $r$ function in our
construction.  But we need those two applications to be
variable-disjoint.  We will need this as otherwise we'd have
interference with some of our claims about sizes of backbones and
which variables are fixed and how many variables we have.  These are
requirements not present in any previous work that used the $r$
function of Galil-Cook.  We also will want to be able to have some
literal names (in particular, ``z''-using literal names of the form
$z_\ell$, $z'_\ell$ 
$\overline{z_\ell}$, or $\overline{z'_\ell}$, for
all~$\ell$) available to us that we know are not part of the output
of any application of the Galil-Cook $r$ function; we need them as our
construction involves not just two applications of the $r$ function
but also some additional variables.  We can accomplish all the special
requirements just mentioned as follows.  We will, w.l.o.g.,~assume
that in the output of the Galil-Cook function
$r_{\textit{Galil-Cook}}(N_i,x)$, every variable is of the form $x_j$
(the $x$ there is not a generic example of a letter, but really means
the letter ``x'' just as ``z'' earlier really means the letter ``z''),
where $j$ itself, when viewed as a pair of integers via the standard
fixed correspondence between $\calz^+$ and $\calz^+ \times \calz^+$,
has $N_i$ as its first component or actually, to be completely
precise, the natural number corresponding to $N_i$ in the standard
fixed correspondence between positive integers and strings.  Though
not all implementations of the Galil-Cook $r$ function need have this
property (and in fact, none has previously 
satisfied 
it as far as we know), 
we claim that
one can implement a legal Galil-Cook $r$
function in such a way that it 
has this property yet 
still has the property 
that this $r$ function will have a polynomial-time inversion
function $s$ satisfying the behavior for $s$ mentioned above.
(For those wanting more information on how such a function
$r_{\textit{Galil-Cook}}(N_i,x)$ can be implemented that has 
all the properties claimed above, 
we have included  
as a technical appendix,
a detailed construction we have built that accomplishes this.)

In this paper, as is standard, 
for any machine $N$ we will use $L(N)$ to denote the
language accepted by machine $N$, i.e., the set of strings accepted
by machine $N$.

We now can specify the sets $A$ needed by
Theorems~\ref{t:hard-value-size-k} and~\ref{t:hard-backbone}.
Recall we have (thanks to the assumptions of the theorems) 
fixed a set $B \in (\npcapconp)-\p$.  
$B \in \np$ so let $i$ be a positive integer
such that $N_i$ is a machine from the abovementioned
standard enumeration
such that $L(N_i) = B$.  
$\overline{B} \in \np$ so let $j$ be a positive integer
such that $N_j$ is a machine from the abovementioned
standard enumeration,
such that $L(N_j) = \overline{B}$.
Fix any positive integer $k$.  
Then for the case of that fixed value $k$, the set 
$A$ of 
Theorem~\ref{t:hard-value-size-k} is as follows:
$A_{3,k} = \{   
\left( z_1 \land z_2 \land \cdots \land z_k 
\land (r_{\textit{Galil-Cook}}(N_i,x))\right)
\lor
\left( \overline{z_1} \land \overline{z_2} 
\land \cdots \land \overline{z_k} \land (r_{\textit{Galil-Cook}}(N_j,x))\right)
\condition    x \in \Sigma^*   \}$.
One must keep in mind in what follows that, as per the previous paragraph,
$r_{\textit{Galil-Cook}}$ never outputs literals 
with names involving 
subscripted 
$z$s or $z'$s and the outputs of 
$r_{\textit{Galil-Cook}}(N_i,x)$ and 
$r_{\textit{Galil-Cook}}(N_j,x)$ share no variable names (since $i\neq j$).

Let us argue that $A_{3,k}$ indeed satisfies the requirements of the
$A$ for the ``$k$'' case of Theorem~\ref{t:hard-value-size-k}.  

{\bf\boldmath
$A\in \p$:} Given a string $y$ whose membership in $A$ we are testing,
we make sure $y$ syntactically matches the form of the elements of
$A$ (i.e., elements 
of $A_{3,k}$).  If it does, we then check that its $k$ matches our $k$,
and we use $s$ to get decoded pairs $(i',x')$ and $(j'', x'')$ from
the places in our parsing of $y$ 
where we have formulas---call them $F_{\textit{left}}$ and
$F_{\textit{right}}$---that we are hoping are the outputs of the $r$
function.  
That is, if our input parses as 
$\left( z_1 \land z_2 \land \cdots \land z_k \land (F_{\textit{left}})\right)
\lor
\left( \overline{z_1} \land \overline{z_2} \land 
\cdots \land \overline{z_k} \land (F_{\textit{right}})\right)$, 
then if $s(F_{\textit{left}})$ gives $(N_{i'},x')$ our decoded 
pair is $(i',x')$, and 
$F_{\textit{right}}$ is handled analogously.
We also check to make sure that $x'=x''$, $i=i'$, and
$j=j''$.  If anything mentioned so far fails, then $y\not\in A$.
Otherwise, we check to make sure that
$r_{\textit{Galil-Cook}}(N_i,x') = F_{\textit{left}}$ and
$r_{\textit{Galil-Cook}}(N_j,x') = F_{\textit{right}}$, and reject 
if either equality fails to hold.  (Those checks
are \emph{not} superfluous.  $s$ by definition has to correctly invert
on strings that are the true outputs of 
$r_{\textit{Galil-Cook}}$,
but we did not assume
that $s$ might not output sneaky garbage when given other input
values, and since 
$F_{\textit{left}}$ and $F_{\textit{right}}$ are 
coming from our arbitrary input $y$, they could be
anything.  However, the check we just made defangs the danger just 
mentioned.)  If we have reached this point, we indeed have determined
that $y \in A$, and for each $y \in A$ we will successfully reach this point.

{\bf\boldmath
$A\subseteq \sat$:} For each $x$, either $x \in B$ or $x \not\in B$.
In the former case ($x \in B$), 
$r_{\textit{Galil-Cook}}(N_i,x) \in \sat$ and so 
the left disjunct of 
$\left( z_1 \land z_2 \land \cdots \land z_k 
\land (r_{\textit{Galil-Cook}}(N_i,x))\right)
\lor
\left( \overline{z_1} \land \overline{z_2} 
\land \cdots \land \overline{z_k} 
\land (r_{\textit{Galil-Cook}}(N_j,x))\right)$
can be made true using that satisfying assignment and setting each 
$z_\ell$ to true.
On the other hand, if $x \not\in B$, then 
$r_{\textit{Galil-Cook}}(N_j,x) \in \sat$ and so the whole formula 
can be made true using that satisfying assignment and setting each 
$z_\ell$ to false.

{\bf\boldmath
There is a polynomial-time computable function $f$ such that 
$(\forall F\in A)[f(F)$ outputs a nontrivial backbone of $F]$:}
On input $F \in A$,
$f$ will simply output $\{z_1,z_2,\ldots,z_k\}$, which
is a nontrivial backbone of $F$.  Why is it a nontrivial backbone?
If the $x$ embedded in $F$ satisfies $x\in B$,
then not only does 
$r_{\textit{Galil-Cook}}(N_i,x) \in \sat$ hold, but also 
$r_{\textit{Galil-Cook}}(N_j,x) \not\in\sat$ must hold (since 
otherwise we would have $x \not\in B \land x \in B$, an impossibility).  
So 
if the $x$ embedded in $F$ satisfies $x\in B$, then 
there are satisfying assignments of 
$\left( z_1 \land z_2 \land \cdots \land z_k 
\land (r_{\textit{Galil-Cook}}(N_i,x))\right)
\lor
\left( \overline{z_1} \land \overline{z_2} \land \cdots \land \overline{z_k} 
\land (r_{\textit{Galil-Cook}}(N_j,x))\right)$,
and every one of them has each $z_\ell$ set to true.
Similarly, if 
the $x$ embedded in $F$ satisfies $x\not\in B$, then 
our long formula has satisfying assignments, and 
every one of them has each $z_\ell$ set to false.  Thus 
$\{z_1,z_2,\ldots,z_k\}$ indeed is a size-$k$ backbone.

{\bf\boldmath
There does not exist any polynomial-time computable function $g$ such
that $g(F)$ computes the value of backbone $f(F)$:}
Suppose by way of contradiction that such a polynomial-time 
computable function $g$ does exist.  Then we would have that 
$B \in \p$, by the following algorithm.  Let $f$ be the function 
constructed in the previous paragraph, i.e., the one that 
outputs 
$\{z_1,z_2,\ldots,z_k\}$ when $F \in A$.
Given $x$, in 
polynomial time---$g$ and $f$ are polynomial-time computable,
and although $r$ in general is not since its running time's polynomial
degree varies with its first argument and so is not uniformly
polynomial, $r$ here is used only for the first-component
values $N_i$ and $N_j$ and under that restriction it indeed 
is polynomial-time computable---compute
$g(f(\left( z_1 \land z_2 \land \cdots \land z_k 
\land (r_{\textit{Galil-Cook}}(N_i,x))\right)
\lor
\left( \overline{z_1} \land \overline{z_2} \land \cdots \land \overline{z_k} 
\land (r_{\textit{Galil-Cook}}(N_j,x))\right)))$.
This must either tell us that the $z_\ell$s are true in all satisfying 
assignments, which tells us that it is the left disjunct that is
satisfiable and thus $x\in B$, or it will tell us that the 
$z_\ell$s are false in all satisfying assignments, from which we 
similarly can correctly conclude that $x\not\in B$.  So $B\in \p$, yet
we chose $B$ so as to satisfy $B \in (\npcapconp) - \p$.
Thus our assumption that such a $g$ exists is contradicted.

That ends our proof of 
Theorem~\ref{t:hard-value-size-k}---and so implicitly also of 
Theorem~\ref{t:hard-value}, since 
Theorem~\ref{t:hard-value} follows immediately from 
Theorem~\ref{t:hard-value-size-k}.

Having seen the above proof, the reader will not need a detailed treatment
of the proof of Theorem~\ref{t:hard-backbone}.  Rather, we describe how 
to convert the above construction into one that proves 
Theorem~\ref{t:hard-backbone}.  Recall that for 
the ``$k$'' case of Theorem~\ref{t:hard-value-size-k} our set $A$ was
$\{   
\left( z_1 \land z_2 \land \cdots \land z_k 
\land (r_{\textit{Galil-Cook}}(N_i,x))\right)
\lor
\left( \overline{z_1} \land \overline{z_2} \land \cdots \land \overline{z_k} 
\land (r_{\textit{Galil-Cook}}(N_j,x))\right)
\condition    x \in \Sigma^*   \}$.

For Theorem~\ref{t:hard-backbone}, let us use almost the same set.
Except we will make two types of changes.  First, in the above,
replace the two occurrences of $k$ each with the smallest positive integer 
$m'$ satisfying
${m' \over 
{
\numvars(r_{\textit{Galil-Cook}}(N_i,x))
+
\numvars(r_{\textit{Galil-Cook}}(N_j,x)) 
+ 
2m'
}}
\geq {49 \over 100}$,
where $\numvars$ counts the number of variables in 
a formula, e.g.,
$\numvars(\overline{x_1} \land x_2 \land \overline{x_2}) = 2$,
due to the variables $x_1$ and $x_2$.  Let $m$ henceforward 
denote that value, i.e., the 
smallest (positive integer) $m'$ that satisfies the above equation.
Second, in the right disjunct, change each $\overline{z_\ell}$ to
$\overline{z'_\ell}$.

Note that 
if $x\in B$, then 
$\{z_1 , z_2 , \ldots , z_m\}$ is a backbone whose value is 
the assignment of true to each variable, and that contains at least 49\%
of the variables in the formula that 
$x$ put into $A$.
Similarly, 
if $x\not\in B$, then 
$\{z'_1 , z'_2 , \ldots , z'_m\}$ is a backbone whose value is 
the assignment of false to each variable, and that contains at least 49\%
of the variables in the formula that 
$x$ put into $A$.  
It also is straightforward to see that our thus-created set $A$ 
belong to $\p$ and satisfies $A \subseteq \sat$.

So the only condition of Theorem~\ref{t:hard-backbone} that we still need to
show holds is the claim that, for the just-described $A$, there does
not exist any polynomial-time computable function $g$ such that, on
each $F \in A$, $g(F)$ outputs a backbone whose size is at least 2\%
of $F$'s variables.  Suppose by way of contradiction that such a
function $g$ does exist. We claim that would yield a polynomial-time
algorithm for $B$, contradicting the assumption that $B \not\in \p$.
Let us give such a polynomial-time algorithm.  To test whether
$x \in B$, in polynomial time we create the formula in $A$ that is put
there by $x$, and we run our postulated polynomial-time $g$ on that
formula, and thus we get a backbone, call it $S$, that contains at
least 2\% of $F$'s variables.  Note that we ourselves do not get to
choose which large backbone $g$ outputs, so we must be careful as to
what we assume about the output backbone.  We in particular 
certainly cannot
assume that $g$ happens to always output
either $\{z_1 , z_2 , \ldots , z_m\}$ or $\{z'_1 , z'_2 , \ldots , z'_m\}$.
But we don't need it to.  Note that the two backbones just mentioned 
are variable-disjoint, and each contains 49\% of $F$'s variables.  

Now, there are two cases.  One case is that $S$ 
contains at least one
variable of the form $z_\ell$ or $z'_\ell$.  In that case we are done.
If it contains at least one variable of the form $z_\ell$ then
$x\in B$.
Why?  If $x\in B$, then the left-hand
disjunct of the formula $x$ puts into $A$ is satisfiable and the
right-hand disjunct is not.  From the form of the formula, it is clear
that each $z_\ell$ is always true in each satisfying assignment in
this case, yet that for each $z'_\ell$ there are satisfying
assignments where $z'_\ell$ is true and there are satisfying
assignments where $z'_\ell$ is false.  So if $x\in B$, no $z'_\ell$ can
belong to any backbone.  

By analogous reasoning, if $S$ contains at least one variable of the
form $z'_\ell$ then $x\not\in B$.  (It follows from this and the above
that $S$ cannot possibly contain at least one variable that is a
subscripted $z$ and at least one variable that is a subscripted $z'$,
since then $x$ would have to simultaneously belong and not belong to
$B$.)

The final case to consider is the one in which $S$ does not contain at
least one variable of the form $z_\ell$ or $z'_\ell$.  We argue 
that this case cannot happen.
If this were to happen, then every variable of $F$
other than the variables
$\{z_1 , z_2 , \ldots , z_m,z'_1 , z'_2 , \ldots , z'_m\}$ must be
part of the backbone, since $S$ must involve 2\% of the variables and
$\{z_1 , z_2 , \ldots , z_m,z'_1 , z'_2 , \ldots , z'_m\}$ comprise
98\% of the variables.  But that is impossible.  
We know that the variables used in 
${r}_{\textit{Galil-Cook}}(N_i,x)$ and 
${r}_{\textit{Galil-Cook}}(N_j,x)$ are disjoint.
So the variables in
the one of those two 
that is not the one that is satisfiable can and do take on any value 
in some satisfying assignment, and so cannot be part of any backbone.
(The only remaining worry is the case where one of 
${r}_{\textit{Galil-Cook}}(N_i,x)$ or 
${r}_{\textit{Galil-Cook}}(N_j,x)$ contains no variables.
However, the empty formula is by convention considered 
illegal, in cases such as here where the formulas are not 
considered to be trapped into DNF or CNF\@.  There is a special 
convention regarding empty DNF and CNF formulas, but that is not 
relevant here.)  We have thus concluded the proof of 
Theorem~\ref{t:hard-backbone}.
\end{proof}

The 49\% and 2\% used in
Theorem~\ref{t:hard-backbone}'s statement and proof
are not at all magic, but are just for
concreteness.
It is
immediately clear that our above proof that establishes 
Theorem~\ref{t:hard-backbone}
is in fact tacitly (namely, if one changes---from the fixed constants we
used in that proof---to instead the more flexible constants below)
establishing the following more
general claim (note: the smaller the $\epsilon$ the stronger 
the claim, and so the fact that below we
speak only of $\epsilon \leq 1$
is not a weakness), of which the above theorem is the $\epsilon = 1$ case.

\begin{theorem}\label{t:hard-backbone-generic-case}
For each fixed $\epsilon$, $0 < \epsilon \leq 1$, the following
claim holds.
If $\p\neq \npintconp$, then there exists a set $A \in \p$, 
$A \subseteq \sat$, of 
boolean formulas (each having at least one variable) such that:
\begin{enumerate}
\item Each formula $F\in A$ has a backbone whose size is at 
least $(50 - \epsilon)$\% of $F$'s total number of variables.
\item 
There does not exist any polynomial-time computable function $g$ such
that, on each $F \in A$, $g(F)$ outputs a backbone whose size is 
at least $(2\epsilon)$\% of $F$'s variables.
\end{enumerate}
\end{theorem}

\subsection{Frequency of Hardness}\label{ss:frequency}
A practical person might worry about 
results of the previous section
in the following way. (Here, 
$|F|$ will denote the number of bits in the representation of $F$.)
``Just
because something is hard, doesn't mean it is hard \emph{often}.  For
example, consider Theorem~\ref{t:hard-backbone}.  Perhaps there is a
polynomial-time function $g'$ that, though it on infinitely many
$F\in A$ fails to compute the value of the backbone $f(F)$,
has the property that 
for each $F \in A$ for which it fails it then is correct on the 
(in
lexicographical order) 
next 
$2^{2^{2^{2^{|F|}}}}$ elements of $A$.
In
this case, the theorem is indeed true, but it is a worst-case extreme
that doesn't recognize that in reality the errors may be few and
far---very, very far---between.''

In this section, we address that reasonable worry.  We show that if
\emph{even one problem} in $\npintconp$ is frequently hard, then the
sets in our previous sections can be made ``almost'' as frequently
hard, in a sense of ``almost'' that we will make formal and specific.
Since it is generally believed---for example due to the 
generally believed typical-case hardness of integer factoring---that 
there are sets in $\npintconp$
that are 
quite 
frequently hard, 
it follows that the $2^{2^{2^{2^{|F|}}}}$ behavior
our practical skeptic was speculating about cannot happen.  Or at
least, if that behavior did happen, then that would imply that every single
problem in $\npintconp$ has polynomial-time heuristic algorithms that
make extraordinarily few errors.

Note that no one currently knows for sure how frequently-hard 
problems in $\npintconp$ can be.  But our results are showing that,
whatever that frequency is, sets of the sort we've been constructing
are hard ``almost'' as frequently.  This can at first seem a bit of a
strange notion to get one's head around, especially 
as complexity theory often doesn't pay much attention to 
frequency-of-hardness issues (though such issues in complexity 
theory can be traced 
back at least as far as the work of 
Sch\"oning~\cite{sch:j:closeness}).  But this actually is
analogous to something every computer science researcher knows
well, namely, NP-completeness.  No one today knows for sure whether
any NP problems are not in P\@.  But despite that the longstanding
NP-completeness framework lets one right now, today, prove clearly for
specific problems that \emph{if} any NP problem is not in P then that
specific problem is not in P\@.  The results of this section are about
an analogous type of argument, except regarding frequency of hardness.

We now give our frequency-of-hardness version of Theorem~\ref{t:hard-value}.
A claim is said to hold \emph{for almost every} $n$ if there exists
an $n_0$ beyond which the claims always holds, i.e., the claim fails at 
most at a finite number of values of $n$.  (In the theorems of this section,
$n$'s universe is the natural numbers, $\{0,1,2,\ldots\}$.   And we will
defer the proving of this section's theorems until the end of this section,
where we will argue that the results in effect follow from the constructions
of the previous section.)

\begin{theorem}\label{t:frequency-hard-value}
  If $h$ is any nondecreasing function and for some $B \in \npcapconp$
  it holds that each polynomial-time algorithm, viewed as a heuristic 
  algorithm for testing membership in $B$, 
  for almost every $n$ (respectively, for infinitely many $n$)
  errs on at least $h(n)$ of the strings whose length is at most $n$,
  then there exist an $\epsilon > 0$ and a set $A\in \p$, $A\subseteq \sat$,
  of boolean formulas such that:
\begin{enumerate}
\item There is a polynomial-time computable function $f$ such that 
$(\forall F\in A)[f(F)$ outputs a nontrivial backbone of $F]$.
\item Each polynomial-time computable function $g$ will 
err (i.e., will fail to compute the value of backbone $f(F)$),
for almost every  $n$ (respectively, for infinitely many $n$),
on at least $h(n^\epsilon)$ of the strings in $A$ of length at most $n$.
\end{enumerate}
\end{theorem}

The precisely analogous result holds for
Theorem~\ref{t:hard-value-size-k}.  The analogous results
also hold for Theorems~\ref{t:hard-backbone}
and~\ref{t:hard-backbone-generic-case}, and to be explicit, we state that 
for Theorem~\ref{t:hard-backbone}
as the following theorem (and from this the analogue for
Theorem~\ref{t:hard-backbone-generic-case} will be implicitly clear).

\begin{theorem}\label{t:frequency-hard-backbone}
  If $h$ is any nondecreasing function and for some $B \in \npcapconp$
  it holds that each polynomial-time algorithm, viewed as a heuristic 
  algorithm for testing membership in $B$, 
  for almost every $n$ (respectively, for infinitely many $n$)
  errs on at least $h(n)$ of the strings whose length is at most $n$,
  then there exist an $\epsilon > 0$ and a set $A\in \p$, $A\subseteq \sat$,
  of boolean formulas such that:
\begin{enumerate}
\item Each formula $F\in A$ has a backbone whose size is at 
least 49\% of $F$'s total number of variables.
\item Each polynomial-time computable function $g$ will 
err (i.e., will fail to compute a set of size at least 2\%
of $F$'s variables that is a backbone of $F$),
for almost every  $n$ (respectively, for infinitely many $n$),
on at least $h(n^\epsilon)$ of the strings in $A$ of length at most $n$.
\end{enumerate}
\end{theorem}

What the above theorems say, looking at the contrapositives to the
above results, is that if any of our above cases have polynomial-time
heuristic algorithms that don't make errors too frequently, then
\emph{every single set in $\npcapconp$} (even those related to integer
factoring) has polynomial-time heuristic algorithms that don't make
errors too frequently.

To make the meaning of the above results clearer, and to be completely
open with our readers, it is important to have a frank discussion
about the effect of the ``$\epsilon$'' in the above results.  Let us
do this in two steps.  First, we give as concrete examples two
central types of growth rates that fall between polynomial and
exponential.  And second, we discuss how innocuous or
noninnocuous the ``$\epsilon$'' above is.

As to our examples, suppose that for some fixed $c>0$ a particular
function $h(n)$ satisfies $h(n) = 2^{\omega((\log n)^c )}$.  Note that
for each fixed $\epsilon > 0$, it hold that the function $h'(n)$
defined as $h(n^\epsilon)$ itself satisfies the same bound,
$h'(n) = 2^{\omega((\log n)^c )}$.  (Of course, the constant implicit
in the ``$\omega$'' potentially has become smaller in the latter
case.)  
Similarly, suppose that a particular
function $h(n)$ satisfies $h(n) = 2^{n^{\omega(1)}}$.  Then for each
fixed $\epsilon > 0$ it will hold that 
$h(n^{\epsilon}) = 2^{n^{\omega(1)}}$.

The above at a casual glance might suggest that the weakening of the
frequency claims between the most frequently hard problems in
$\npcapconp$ and our problems is a ``mere'' changing of a constant.
In some sense it is, but constants that are standing on the shoulders
of exponents have more of a kick than constants sitting on the ground
floor.  And so as a practical matter, the difference in the actual
numbers when one substitutes in for them can be 
large.  On the other hand,
polynomial-time reductions sit at the heart of computer science's
formalization of its problems, and density distortions from $n$ to
$n^\epsilon$ based on the stretching of reductions are simply inherent
in the standard approaches of theory, since those are the distortions
one gets due to polynomial-time reductions being able to stretch
their inputs to length $n^{1/\epsilon}$, i.e., polynomially.  For
example, it is well known that if $B$ is an NP-complete set, then for
every $\epsilon > 0$ it hold that $B$ is polynomial-time isomorphic
(which is theoretical computer science's strongest standard notion of
them being ``essentially the same problem'') to some set $B'$ that
contains at most $2^{n^\epsilon}$ strings at each length $n$.

Simply put, the ``almost'' in our ``almost as frequent'' claims is the
natural, strong claim, judged by the amounts of slack that in
theoretical computer are considered innocuous.  And the results do
give insight into how much the density does or does not change, e.g.,
the first above example shows that quasi-polynomial lower bounds on
error frequency remain quasi-polynomial lower bounds on error
frequency.  However, on the other hand, there is a weakening, and even
though it is in a ``constant,'' that constant is 
in an 
exponent and so 
can 
alter the numerical frequency
quite a 
bit.\footnote{%
\label{f:epsilon-vs-time}In reality, 
how nastily small will the ``$\epsilon$'' be?
{}From looking inside the proofs of the results and thinking hard
about the lengths of the formulas involved in the proofs (and
resulting from the ``Galil'' version of Cook-Karp-Levin's Theorem 
that we will be
discussing in the next section), one can see that $\epsilon$'s value 
is primarily controlled by the running time of the NP machines
for the NP sets $B$ and $\overline{B}$ from the theorems 
of this section.  If those machines run in time $\bigo(n^q)$, then
$\epsilon$ will 
vary, viewed as a function of $q$, roughly as (some constant 
times) the inverse of $q$.
For the particular case of
Theorem~\ref{t:frequency-hard-value}, for example, our coming proof
will make it clear that, where $q$ is as just stated, for any
$\delta > 0$ one can make the $\epsilon$ in the theorem's
$h(n^\epsilon)$ be $\frac{1}{q} - \delta$, i.e., the theorem's
lower bound
is 
$h(n^{\frac{1}{q} - \delta})$. Indeed, the penultimate paragraph of
that proof will actually establish---in the text immediately
before it simplifies by introducing $\delta$---a slightly
stronger bound, namely,
that there will be a constant $c>0$
(related to not just
the degree of but also the multiplicative constant of
the running-time
bound on the NP machines for $B$ and $\overline{B}$)
such that the
lower bound the theorem is speaking of can be taken as 
$h(c \, n^{\frac{1}{q}})$.}

Our use of $B$ and $A$ here reflects that of the theorems
in this section.
The crucial thing to note is that 
the mapping from strings $x$ (as to whether they belong to $B$) 
into the string that $x$ puts into $A$ is (a)~polynomial-time computable
(and so the one string that $x$ puts into $A$ is at most polynomially
longer than $x$), and (b) one-to-one.  

So any collection of $m$ instances up to a given length $n$ that fool
a particular polynomial-time algorithm for $B$ are associated with at
least $m$ distinct instances in $A$ all of length at most $n^q$
(%
where the
polynomial bound on the length of the formula that $x$,
$|x| = n$, puts into $A$
is that its length is $n^q$ or less%
\footnote{%
  We have for simplicity
  left out any
lower-order terms and the 
leading-term constant,
but that is legal except at $n \in \{0,1\}$---since
starting with $n=2$ we can 
boost $q$ 
if needed---and no finite
set of values, such as $\{0,1\}$ can cause problems to our theorem, as
it is about the ``infinitely-often'' and ``almost-everywhere'' cases.
However, such boosting does potentially interfere with the 
inverse-of-$k$ relation mentioned in Footnote~\ref{f:epsilon-vs-time}, 
and so 
if we wanted to maintain that, we would 
in this argument instead use a lowest-degree-possible monotonic 
polynomial bounding the growth rate.}%
).
So if one had an algorithm for the ``$A$'' set such that the algorithm
had at most $m'$ errors on the strings up to length $n^q$, it would
certainly imply an algorithm for $B$ that up to length $n^{1/q}$ made
at most $m'$ errors.  Namely, one's heuristic of that form for $B$ 
would be to take $x$, map it to 
the string it put into $A$, and then run the 
heuristic for $A$ on that string.

The results of this section are
the immediate
consequences of
this observation, applied to
the constructions/results of the previous section.
To make completely clear that that is the case and why it is the case,
we now provide a more detailed explanation of the proof of one of this
section's theorems, namely, Theorem~\ref{t:frequency-hard-value}.

\begin{proof}[Proof of  Theorem~\ref{t:frequency-hard-value}]
Let $A$ be defined as $A_{3,1}$ in Section~\ref{ss:basic}, i.e.: 
$$A = A_{3,1}= \{   
\left( z_1
\land (r_{\textit{Galil-Cook}}(N_i,x))\right)
\lor
\left( \overline{z_1} \land (r_{\textit{Galil-Cook}}(N_j,x))\right)
\condition    x \in \Sigma^*   \}$$
where $N_i$ and $N_j$ are Turing machines such that $L(N_i)=B$ and
$L(N_j)=\overline{B}$. We know from our discussion in
Section~\ref{ss:basic} that $A\in\p$ and that, given any formula $F\in
A$, $\{z_1\}$ is a nontrivial backbone of that formula, thus the
function $f(F)=\{z_1\}$ satisfies the requirements of the first part
of Theorem~\ref{t:frequency-hard-value}.

Since $r_\textit{Galil-Cook}$ runs in polynomial time, there exist
polynomials $p_i$ and $p_j$ of equal degree $q$ such that the length
of the formula $r_\textit{Galil-Cook}(N_i,x)$ is at most $p_i(|x|)$
and the length of the formula $r_\textit{Galil-Cook}(N_j,x)$ is at
most $p_j(|x|)$. Note that there will exist natural numbers $k$ and
$N$ such that for all $n> N$, $k \, n^q\geq
|\left( z_1 \land
(r_{\textit{Galil-Cook}}(N_i,x))\right) \lor \left( \overline{z_1}
\land (r_{\textit{Galil-Cook}}(N_j,x))\right)|$ for all strings $x$
whose length is at most $n$.
Let $n_B>N$ be a natural number such that
every polynomial-time algorithm, viewed as a heuristic for testing
membership in $B$, errs on at least $h(n_B)$ of the strings whose
length is at most $n_B$.
We claim that every polynomial-time algorithm, viewed as a heuristic
for computing the value of $f(F)=\{z_1\}$ for
inputs $F\in A$, errs on at
least $h(n_B)$ of the strings whose length is at most
$k\,
(n_B)^q$.  Let us define $n_A$ by 
$n_A=k\,(n_B)^q$.  
Making that variable substitution in our claim, we have
that every polynomial-time algorithm, viewed as a heuristic for
computing the value of $f(F)=\{z_1\}$ for inputs $F\in A$, errs on at
least $h\left(k^{-\frac{1}{q}}\,(n_A)^\frac{1}{q}\right)$ of the strings
whose length is at most $n_A$.
For any $\delta>0$ it certainly holds that, for almost all $n$,
$k^{-\frac{1}{q}}\,n^{\frac{1}{q}} \geq n^{\frac{1}{q}-\delta}$, and
thus, since $h$ is nondecreasing, it certainly
holds that, for almost all $n$,
$h\left(k^{-\frac{1}{q}}\,n^{\frac{1}{q}}\right) \geq
h\left(n^{\frac{1}{q}-\delta}\right)$.
Depending on which part of the 
``respectively'' in the theorem's statement one is speaking of,
from the assumptions 
we have that almost every $n>N$ can
take the role of $n_B$ (respectively, infinitely many $n>N$ can take
the role of $n_B$). Thus setting $\epsilon=\frac{1}{q}-\delta$ proves
that $A$ satisfies the second part of
Theorem~\ref{t:frequency-hard-value}.

To prove our claim, notice that, by our choice of $n_B$,
for all inputs $x$ of length at most
$n_B$ the length of $\left( z_1 \land
(r_{\textit{Galil-Cook}}(N_i,x))\right) \lor \left( \overline{z_1}
\land (r_{\textit{Galil-Cook}}(N_j,x))\right)$ is at most $k \, (n_B)^q$.
Assume, by contradiction, that there
exists a polynomial-time algorithm $g'$ that, viewed as a heuristic
for computing the value of $f(F)=\{z_1\}$ for inputs $F\in A$, errs on
fewer than $h(n_B)$ of the strings of length at most
$k\,(n_B)^q$. Consider the following polynomial-time heuristic for
testing membership in $B$: 
On input $x$, calculate $v=g'(\left( z_1
\land (r_{\textit{Galil-Cook}}(N_i,x))\right) \lor \left(
\overline{z_1} \land (r_{\textit{Galil-Cook}}(N_j,x))\right))$;
if $v$ sets $z_1$ to true then output $x\in B$ and if $v$ sets $z_1$
to false output $x\notin B$.
Based on our discussion in
Section~\ref{ss:basic}, this heuristic will err exactly when $g'$ errs
since, for instance, if $v$ sets $z_1$ to true
but the correct value sets $z_1$ to false that would
imply $x\notin B$. But $g'$ errs on fewer than $h(n_B)$ inputs of
length at most $k\,n_B^q$, so the polynomial-time heuristic we
just constructed errs on fewer than $h(n_B)$ inputs $x$ of length at
most $n_B$, a contradiction.
\end{proof}

\section{Related Work}\label{s:related}
Our results can be viewed as part of a line of work that though
interesting is, unfortunately, so
underpopulated as to barely merit being called a line of work.
The 
true inspiration for
this work was
an insightful
structural complexity theory
paper of Alan Demers and Allan Borodin
\shortcite{bor-dem:t:selfreducibility} from the 1970s, which never
appeared in any form other than as a technical report.
Their
paper
in effect showed sufficient conditions for
creating simple sets of satisfiable formulas such that it was unclear
why they were satisfiable.

Borodin and Demers's work
has been used only very rarely.  In particular, it has been used to get
characterizations regarding unambiguous
computation~\cite{har-hem:j:up}, and Rothe and his collaborators have
used it in various contexts to study the complexity of
certificates~\cite{hem-rot-wec:j:hard-certificates,rot:thesis-habilitation:certificates},
see also Fenner et al.~\shortcite{fen-for-nai-rog:j:inverse}
and Valiant~\shortcite{val:j:checking}.
Also, 
one paper---Hemaspaandra, Hemaspaandra, and
Menton~\shortcite{hem-hem-men:c:search-versus-decision}---shows that
the work has a connection to an ``applied'' area, namely,
that paper shows that some problems
about
the manipulation of elections
have the property that if $\p \neq \npcapconp$
then their search versions are not polynomial-time Turing reducible to
their decision problems---a rare behavior among the most familiar
seemingly hard sets in computer science, since so-called
self-reducibility~\cite{mey-pat:t:int} is known to preclude that
possibility for most standard NP-complete problems.  The key issue
that 2013 paper left open is whether the type of techniques it used,
descended from Borodin and
Demers~\shortcite{bor-dem:t:selfreducibility}, might be relevant
in other domains, or whether its results were a one-shot oddity.
The present paper in effect is arguing that the former is the case.
Backbones are a topic important in both theory
and artificial intelligence.
This paper shows that 
the inspiration of the line of work initiated 
by Borodin and
Demers~\shortcite{bor-dem:t:selfreducibility} 
can be used to 
establish 
the opacity of backbones.
It is important to acknowledge that our proofs regarding
Section~\ref{ss:basic} are drawing on elements of the insights of
Borodin and Demers~\shortcite{bor-dem:t:selfreducibility}, although in 
ways 
unanticipated by that paper.

This paper uses 
density transfer arguments in the context of
Borodin-Demers arguments.  To the best of our knowledge,
the only paper to previously do that is the work---in the quite different
context of computational social choice theory---of
Hemaspaandra, Hemaspaandra, and
Menton~\shortcite{hem-hem-men:c:search-versus-decision}.

Finally, note that most of our results
rely on the assumption that $\pisnotinter$,
which as noted above is likely true, since if it is false then integer
factoring is in $\p$ and the RSA encryption scheme falls.
What results can one obtain if one allows oneself
to assume only $\p \neq \np$?  That question is explored
in a recent paper by
the authors that was motivated by the present
work~\cite{hem-nar:ctoappear:backdoors-opacity}.
However, that later paper's results, due to the weaker assumption, do not
at all address the issues, central to the present paper, of outputting a backbone
or outputting the value of a backbone.

\section{Conclusions}
We argued, under assumptions widely believed to be true such as
the hardness of integer factoring, that knowing a large backbone exists
doesn't mean one can efficiently find a large backbone, and finding a
nontrivial backbone doesn't mean one can efficiently find its value.
Further, we showed that one can ensure that these effects are not very
infrequent, but rather that they can be made to happen with ``almost'' the
same density of occurrence as the error rates of the most densely hard
sets in $\npcapconp$.  

\paragraph*{Acknowledgments}
We are grateful to the anonymous AAAI-17 reviewers for 
their helpful comments and suggestions.

\appendix

\section*{\boldmath{}Appendix: 
Construction of a Galil-Cook 
$r$ Function with the Properties Claimed 
in~Section~\protect\ref{ss:basic}}\label{s:appendix}
For those who wish to be assured that a Galil-Cook ``$r$'' function
can be implemented so as to have all the properties we have 
``without loss of generality'' assumed in 
Section~\protect\ref{ss:basic}, we here provide such an implementation.

Let $N_1$, $N_2$, $\ldots$ be as in 
Section~\protect\ref{ss:basic}.  That is, it is a
fixed, standard enumeration of clocked,
polynomial-time Turing machines, such that 
each $N_i$ runs
within time $n^i+i$ on inputs of length $n$, and 
$N_i$ and $i$ are 
polynomially related in size and easily obtained from each other.
Fix \emph{any} function $r$ that implements the Cook-Karp-Levin reduction.
That is, $r$ is such that
\begin{enumerate}
\item for each $N_i$ and $x$: $x \in L(N_i)$ if and only if $r(N_i,x) \in \sat$.
\item there is a polynomial $p$ such that 
$r(N_i,x)$ runs within time 
polynomial (in particular, with $p$ being the polynomial) in
$|N_i|$ and $|x|^i+i$.
\end{enumerate}
It is very well known that such functions exist. Their existence---the
Cook-Karp-Levin reduction---is proven in almost every textbook that
covers NP-completeness (see, e.g., 
Hopcroft and Ullman~\citeyear{hop-ull:b:automata}),
 and is the key moment that brings the theory
of NP-completeness to life, by transferring the domain from machines
to a concrete problem that itself can be used to show that other concrete
NP problems are themselves NP-complete.

Note that the function $r$ that we have thus fixed is \emph{not}
assumed to necessarily have an ``inversion'' function $s$, and is
\emph{not} assumed to necessarily avoid using literals involving the
letter ``z'', and is \emph{not} assumed to necessarily have the
property that two applications of the $r$ function are guaranteed
to be variable-disjoint if they regard different machines (i.e., are
not both about $N_i$ for the same value $i$).

We now show how to use the above fixed function $r$ as a building
block to build our function $r_{\textit{Galil-Cook}}$, which will have
all the properties just mentioned, yet will retain the time and
reduction-to-SAT properties mentioned above regarding $r$.
$r_{\textit{Galil-Cook}}$, when its first argument is $N_i$ and its
second argument is $x$, does the following.  It simulates the run of
$r$ when its first argument is $N_i$ and its second argument is $x$,
and so computes the formula $r_{\textit{Galil-Cook}}(N_i,x)$, which 
we will henceforth denote by $F$ for conciseness of notation.  It then 
counts the number of variables occurring in that formula $F$; let us denote
that number by $p$.  (So for example 
if the formula $F$ is $z_1 \land \overline{z_1} \land 
\overline{\textit{w}}$, then $p=2$, as there are two variables,
$z_1$ and $\textit{w}$.)  Now, let $F'$ denote $F$, except 
each variable $a$ in $F$ will be replaced in $F'$ 
by the variable $x_{\pairs{N_i,q}}$,
where $q$ is the location of $a$ in lexicographic order within the 
variables of $F$.  
($\pair{\cdot,\cdot}$ is any standard, 
nice,
easily computable,
easily invertible pairing function.)
If we view 
$\textit{w}$ as coming lexicographically before $z_1$, then 
in our example, $F'$ would be 
$x_\pair{N_i,2}
 \land \overline{x_\pair{N_i,2}} \land 
\overline{x_\pair{N_i,1}}$.  
Despite the pairings used, this increases the length of $F$ by at most
a multiplicative factor of the number of bits of $N_i$.  (Since each of 
our uses of 
$r_{\textit{Galil-Cook}}$ in 
Section~\protect\ref{ss:basic} only used the function on 
some two hypothetical, fixed 
machines, the time and length-of-output effect of this 
variable-renaming is at most a multiplicative constant (that depends on 
the two machines), and so is
negligible in standard complexity-analysis terms).
But although using the same trick to encode $x$ into the output by 
pairing it too into the variable names would be valid, it would increase 
the formula size by a multiplicative factor of $|x|$, which is 
not negligible.  So we take a different approach, which instead 
increases the formula size just by an \emph{additive} factor of $\bigo(|x|)$.
$r_{\textit{Galil-Cook}}(N_i,x)$ will output 
$(F') \land (c_0 \lor c_1 \lor c_{b_1} \lor \cdots \lor c_{b_{|x|}})$,
where in the above $b_i$ denotes the value of the $i$th bit of $x$ and 
each $c_0$ above means to write $\overline{x_\pair{N_i,p+1}}$
and each $c_1$ above means to write ${x_\pair{N_i,p+1}}$.

So, in our running example, if the value of $x$ was $101$, 
$r_{\textit{Galil-Cook}}(N_i,x)$ would be 
$(x_\pair{N_i,2}
 \land \overline{x_\pair{N_i,2}} \land 
\overline{x_\pair{N_i,1}}) \land (
\overline{x_\pair{N_i,3}}
\lor 
{x_\pair{N_i,3}}
\lor 
{x_\pair{N_i,3}}
\lor 
\overline{x_\pair{N_i,3}}
\lor 
{x_\pair{N_i,3}})$.  

It is not hard to see that the 
$r_{\textit{Galil-Cook}}$ we have constructed has all the
promised properties.  It has the correct running time, it validly reduces from
whether $N_i$ accepts $x$ to the issue of whether 
$r_{\textit{Galil-Cook}}(N_i,x)$ is in $\sat$, 
it never outputs any literal involving
the letter z,
all its literals in fact are tagged by the $N_i$ in use and so 
two applications created with regard to different machines (e.g.,
$N_4$ and $N_7$) are guaranteed 
to have variable-disjoint outputs, and it even is such that the desired
$s$ function exists.  Our $s$ function will take an input, parse it 
to get $(A) \land (B)$, will decode $N_i$ from the variables names in $A$
and will decode $x$ from the fact that it is basically written out by 
the bits encoded by all but the first two disjuncts of $B$, and then will
output 
the pair $(N_i,x)$.  If anything goes wrong in that process, as to 
unexpected syntax or so on, then what we were given is not an actual
output of some run of 
$r_{\textit{Galil-Cook}}$ on a legal input, and we can output any junk
pair that we like, without violating our promise as to the behavior of $s$.
(Some inputs that are not valid outputs of 
$r_{\textit{Galil-Cook}}$ will not trigger the above ``if anything goes
wrong,'' since we did not here take the $(N_i,x)$ we are about 
to output and compute
$r_{\textit{Galil-Cook}}(N_i,x)$ to see whether the output of that matches our 
input. But we do not need to. The needed behavior here is that all
valid inputs have the right output, and we have achieved that.)

%
\newcommand{\etalchar}[1]{$^{#1}$}

\bibliographystyle{alpha}

\end{document}